# Three-stage intelligent support of clinical decision making for higher trust, validity, and explainability

*Sergey V. Kovalchuk*[1], *Georgy D. Kopanitsa*[1], *Ilia V. Derevitskii*[1], *Daria A. Savitskaya*[2]

[1]ITMO University, Saint Petersburg, Russia

[2]Almazov National Medical Research Centre, Institute of Endocrinology, Saint Petersburg, Russia

kovalchuk@itmo.ru, kopanitsa@itmo.ru, ivderevitckii@itmo.ru, savitskayadaria@gmail.com

**Abstract.** The paper presents a conceptual framework for building practically applicable clinical decision support systems (CDSSs) using data-driven (DD) predictive modelling. With the proposed framework we have tried to fill the gap between experimental CDSS implementations widely covered in the literature and solutions acceptable by physicians in daily practice. The framework is based on a three-stage approach where DD model definition is accomplished with practical norms referencing (scales, clinical recommendations, etc.) and explanation of the prediction results and recommendations. The approach is aimed at increasing the applicability of CDSSs based on DD models through better integration into decision context and higher explainability. The approach has been implemented in software solutions and tested within a case study in type 2 diabetes mellitus (T2DM) prediction, enabling us to improve known clinical scales (such as FINDRISK) while keeping the problem-specific reasoning interface similar to existing applications. A survey was performed to assess and investigate the acceptance level and provide insights on the influences of the introduced framework's element on physicians' behavior.

## 1. Introduction

Clinical decision support systems (CDSSs) have a long history. Starting from the late 50s to the present days, they have been developed through several architectural approaches [1]. Still, having a proven capability of CDSSs to improve clinical practice, their real-world application is limited due to multiple issues, including lack of trust, validity, scalability, etc. Clinical professionals who use CDSSs often over-rely on system suggestions, even if these suggestions are wrong [2]. On the other hand, attention to the recommendations produced by CDSSs varies over time. They can cause an increase in quality of clinical decisions, which is followed by a decrease in quality due to lower attention to the recommendations [3]. At the same time, CDSSs cannot be entirely reliable due to uncertainties and lack of data, and thus the correctness of their outputs may affect the quality of decision-making [4]. Therefore, it is essential that clinical professionals do not trust CDSSs blindly. This should be supported by interoperability of decision support models [5]. Providing explanations could potentially mitigate misplaced trust in the system and over-reliance [6].

The recent development of machine learning and artificial intelligence (AI) technologies [7] emphasizes both the strengths and weaknesses of CDSSs, which were mentioned above. Moreover, relying on available data, such technologies may fall into dependency on data and uncertainty of information. Uncertainties originate in almost every step of the clinical decision-making process.

The systematic and thorough verification and validation of clinical decision support systems before they are released to customers is a crucial aspect of optimal software design [8]. Ignorance of this step could lead to negative consequences of relying on a CDSSs with uncertain outputs. The verification and validation of decision support systems applies various methods to detect problems with CDSSs during testing [9]. This includes, among others, validation of decision support models [10], [11], medical data [12], and clinical workflows [13]. Still, the absence of a commonly accepted approach or methodology for the verification and validation of CDSSs as a specific class of solutions may lead to the limited use of intelligent technologies (see, e.g. [14] for discussion of possible issues of IBM Watson).

Taking the mentioned issues into account, we see a gap between experimental development of CDSS widely covered in scientific literature and real-world application of CDSS. Within the presented study we are trying to fill this gap with the proposed conceptual framework aimed toward enriching DD-models with normative references, which frames the practice of a physician and deeper domain-specific explanation of the prediction results.

The issue of combining predictive power of data-driven models and interpretability of trustable, manually checked, and widely applied in practice rule-based algorithms can potentially be solved if we add validation to a common data-driven modeling pipeline. To be able to efficiently develop such hybrid systems, we propose a framework with three clear stages that represents a process of evolution of the decision support system. Following this framework, a developer will be able to base design and architecture of a decision support system on the formal vocabulary and tools to combine a predictive power of data-driven systems with the trust and interpretability of rule-based methods. It can improve the functionality, integrability, and interpretability of CDSS in the existing clinical business processes (BPs). Conceptual integration of all stages will provide a human-understandable interface for decision support systems.

The remaining part of the paper is structured as follows. Section 2 summarizes some available studies in the area of application of CDSSs and AI in the medical domain. Section 3 introduces the proposed approach based on the identified limitations in direct application of AI technologies in the area. The next section shows a case study and implementation details to assess the applicability of the approach to practical problems. Section 5 provides a discussion on the limitations of the approach and the obtained results. The last section provides a conclusion and describes possible directions of further development.

## 2. Related works

Recent systematic reviews [15][16][17] show that the application of CDSSs in clinical flows is still limited due to fragmented workflows especially in the case of stand-alone implementations. Despite that, continuous CDSSs suffer from semantic interoperability issues. Many CDSSs exist as awkward stand-alone systems, or cannot communicate effectively with other clinical systems [18].

Within the current study, we do not intend to review all the CDSS implementations, but we think it is important to present the most prominent instances that support our and the authors' of the existing systematic reviews conclusions that clinical decision support systems do not provide a combination of explainability, trust and validity.

For example, the clinical decision support system developed by Kirby et al [1] alerts doctors ordering echocardiograms if patient results met criteria based on valve severity. The implementation of the system increased a referral by 24.6% increase in referral rates. However, Medical outcomes were not reported in the study. The authors did not provide an assessment of accuracy or other efficiency metrics of the systems. IT system doesn't provide interpretations of the decisions.

A hybrid CDSS by Caballero-Ruiz et al [2] managed to reduce time devoted by clinicians to patients during face-to-face medical encounters by 89%. The system is based on the hybrid approach with glycaemia values, which were automatically labelled with their associated meal by a classifier based on the Expectation Maximization clustering algorithm and a C4.5 decision tree learning algorithm. Two finite automata are combined to determine the patient's metabolic condition, which is analysed by a rule-based knowledge base to generate therapy adjustment recommendations. The performance of the CDSS was validated within a randomized controlled clinical trial.

A rule-based system developed by Raj et al [3] provides computerized decision support of opiates dosing at point of care for the physician based on the electronic questionnaires completed by the patients. A controlled before-and-after study demonstrated no improvement of the pain intensity, therefore no significant differences in dose of opiates compared with control. The system also had no effect on practitioner performance. This happened despite a descent accuracy of the decision support system. The interpretations were not provided by the CDSS.

A data-driven clinical decision support system, which identifies prescriptions with a high risk of medication error by Corny et al [5], provides precision and recall of 0.81 and 0.75 respectively. It was validated by expert pharmacists. However, it does not provide explanations of the decisions.

Many CDSSs are designed to require the provider to document or source data outside their typical workflows. CDSSs also disrupt clinical processes if they are implemented without consideration of human information processing and behaviors. This is especially critical when a CDSS doesn't provide enough transparency and explanation of the decisions made. In these cases, disrupted workflows can cause elevated cognitive efforts, more time needed to complete clinical tasks, and less time for patients.

CDSSs rely on data from external, dynamic systems, and without proper preprocessing, this can create unprecedented pitfalls [19]. In poorly designed CDSSs, users may start developing workarounds that compromise data, transparency, and explainability of the decisions.

The Barcelona Declaration for the Proper Development and Usage of Artificial Intelligence in Europe [20] divides CDSSs into two fundamentally different types: knowledge-based and data-driven. Knowledge-based methodologies are well established but less able to exploit large volumes of data and often rely on manually developed features and rules [21]–[26]. Validation of knowledge-based models can be performed by an expert review as they are usually based on clinical guidelines. So, ensuring that the rules are correctly defined is sufficient in most of the cases [27], [28]. We think that the distinction that the declaration provides is good for classification, however, clinical practice shows that more and more hybrid systems are being used to support clinical decisions [29]–[31].

Data-driven methods employ large amounts of empirical data processed with statistical machine learning methods to extract patterns [32]–[40]. The need for validity and explainability is currently recognized as an essential problem to be solved by researchers [29]. Validation of data-driven models often requires lengthy and expensive clinical evaluation using metrics that are intuitive to clinicians, go beyond the measures of technical accuracy, and include quality of care and clinical effectiveness [41]. Implementation and adoption of machine learning methods can be reasonably straightforward. Still, the interpretation of the model outcomes is sometimes complicated and indistinct due to the black-box nature of machine learning models [42]. Data-driven model's interpretability is a developing research subject, but not all machine learning approaches lend themselves well to interpretability. Interpretability also usually comes at the cost of performance. In our study we propose an approach that will add interpretability to the data-driven approach preserving its predictive power [43], [44].

The solution can be a combination of knowledge-based and data-driven models and methods. Even though every system should implement its own explanation module, which may vary based on the application domain and the adopted methodologies and algorithms, we believe that, in general, the synergy between different knowledge types and AI methodologies can become a promising strategy to deal with the issues of transparency and explainability.

### 3. Three-stage clinical decision making

#### 3.1 Key objectives and actors

Within our experience in research and development in the area of CDSSs, we often see that the limitations of large-scale CDSS applicability and translation to various diseases often come from the "local" nature of studies with limited connection to the real-life conditions and requirements. During clinical decision making, these conditions include a) requirement to work in alignment with existing official (governmental, organizational) regulations and policies, which are usually expressed in documents and laws; b) requirement to provide a complete explanation and proof of the proposed recommendations and estimations to decision makers. Thus, within this study, we aim at improving the level of integration of data-driven AI solutions into the clinical decision process with explicit interconnection of concepts, rules, policies, etc. We are improving the explanation and validation procedures with provided recommendations for higher trust and applicability.

Aimed towards this goal, we consider a general scenario of CDSS application in clinical practices where a decision maker (usually a physician) is supported by AI-based information during patient treatment process. Here, decisions may concern diagnosis, treatment planning, collecting additional information, etc. where supportive information within a CDSS may include predictions of diagnosis and outcomes, risk assessment, estimation of effectiveness of treatment plans, interpretation of tests and images, and many others. Considering a more general pipeline, the list of actors should be extended with a) a patient whose treatment is under consideration by the physician and b) experts who indirectly participate in decision making by providing clinical recommendations, elaborating policies, issuing standards, etc.

### 3.2 Conceptual framework

This section proposes the basic structure and critical requirements in the proposed framework for CDSS implementation. The framework is based on an approach, which includes three main stages (see Fig. 1). The holistic way of the application of this approach is based on the tight interconnection of all three stages to support the integration of a CDSS into a clinical decision-making pipeline.

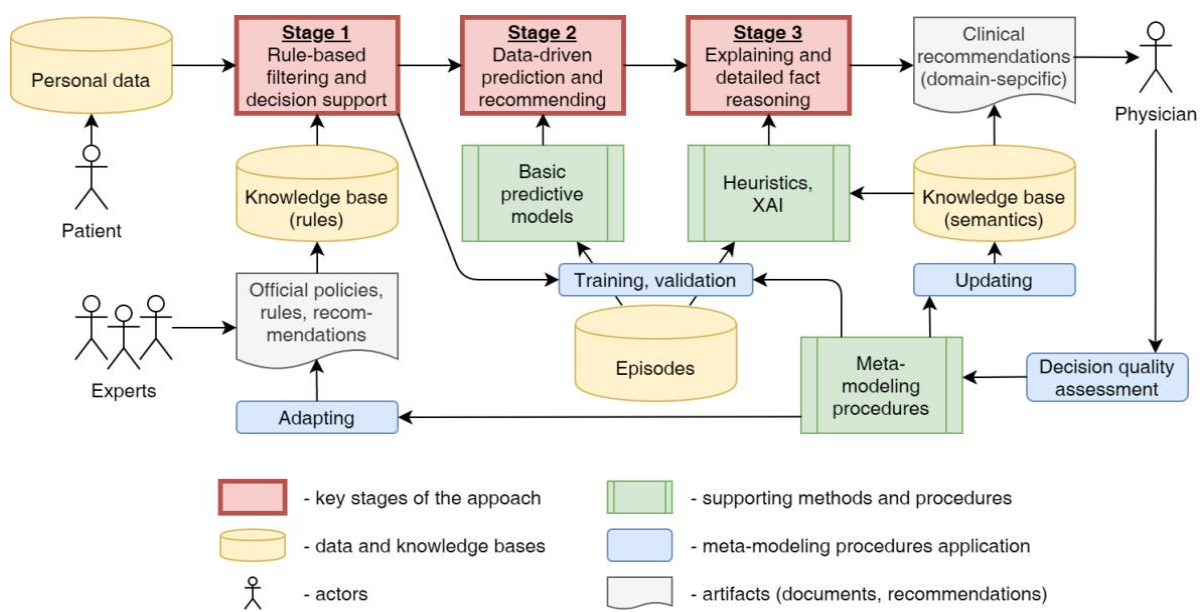

Figure 1 – Three stages of clinical decision support

***Stage 1 – Basic reasoning.*** This stage includes digital implementation of well-grounded and known practices, policies, and rules implemented in conventional healthcare. An essential part of this stage is a direct referencing to existing regulation policies. This type of referencing is necessary to a) get into the context of the existing medical practice; b) define a subset of relevant data and applicable knowledge for a specific disease, patient, case, etc.; c) provide a reliable and trusted foundation for further reasoning. In many cases, the relevant policies are expressed by experts in the form of standards or laws. Thus, to implement this stage, the corresponding artifacts (blueprints, guidelines, etc.) should be structured and formalized. The implementation can be performed in various knowledge models. Common approaches include rule bases, decision trees, and semantic structures (e.g., ontologies).

***Stage 2 – Data-driven predictive modeling.*** The second stage includes the application of data-driven models, including machine learning algorithms. The models should be applied within the context defined at the first stage of the approach. The key result of this stage is the extension of available information on a particular case. A common problem definition in this stage is classification (prediction of diagnosis, complications, outcomes, etc.) or regression (prediction of risks, vital characteristics, probability of events, etc.) task. To address this problem, various data-driven techniques (machine learning approaches, statistics, data assimilation, etc.) are employed. A hybrid implementation of this stage may also be based on including additional approaches to generate synthetic data, assess different scenarios through the simulation, etc.

***Stage 3 – Explaining and detailed fact reasoning.*** This later stage of the application is essential due to several reasons. Firstly, the information set collected during the first two stages has high diversity. This leads to different levels of uncertainty and trust, which is often critical for CDSS applicability. Secondly, explainability (e.g., with technologies like eXplainable Artificial Intelligence, XAI or Interpretable Machine Learning, IML) and domain integration (domain semantics) are vital for performing better human-computer interaction during clinical decision making [45]. Finally, weighting and structuring of obtained results are needed to provide better recommendations [46]. As a result, filtering, integration, and assessment procedures are implemented in this stage using XAI, heuristics, and explainable metamodeling [47].

The framework accomplishes the middle stage (Stage 2, which is widely adopted in many recent studies) with two additional elements important for practical application. Stage 1 arranges the data-driven modeling procedures with existing "regulation field", its limitations and rules. Stage 3 proceeds with the results of data-driven modeling to augment them with domain-specific explanation, validation, and proof. Moreover, the connected stages may form a loop to improve the available data-driven models and even policies being applied in a practical clinical decision problem by a domain decision maker (physician).

To form a loop, the framework includes a set of metamodeling procedures, which evaluate the decision (with possible assessment by a decision maker) and improve three stages of the framework. The key element of the loop is assessing the quality of the decision being made by a decision maker. With this element, the loop can be considered as a tool for global quality assurance implemented through continuous models and knowledge improvement. The metamodeling procedures manage knowledge in various forms within three primary operations (forming a loop for building adaptive intelligent systems):

− identification, training, and validation of basic predictive models and explaining metamodels using historic and actual data;
− updating and structuring domain (semantic) knowledge bases;
− adapting and modification of official policies and recommendations.

The last one should be mentioned separately. We consider this as an essential strategic operation where the results obtained using intelligent technologies can be introduced into the policies either implicitly during policy development, or explicitly considering intelligent technologies as recommended elements. We believe that the current development of AI may lead to such an explicit presence of intelligent technologies in the official policies soon.

### 3.3 Technological background

The mentioned stages may be implemented in various ways employing different technologies. Table 1 summarizes the key features of the stages. Each stage has its own artifacts (models, knowledge bases, etc.), which are to be implemented using various technologies and approaches. This may include technologies mainly from AI or machine learning (ML) to support comprehensive decision making, data processing technologies (including technologies from the area of Big Data), and metamodeling procedures. Selection of an appropriate technology for a particular solution depends on available data and knowledge, as well as on the problem to be solved. Still the key answers to be given by the technology being applied are a) limiting the search area and providing basic conceptual elements (Stage 1); b) proper and well-grounded application of data-driven models (Stage 2); c) appropriate supply of the provided recommendations with proofs and explanations for the decision maker (Stage 3). The summary provided in the table can be used as a starting point to select an appropriate technology regarding the problem and the scenarios of the task.

Table 1 – Summary of technological implementation in three stages

|  | Stage 1 | Stage 2 | Stage 3 |
|---|---|---|---|
| **Preparation of artifacts** | Knowledge acquisition | Training data-driven models | Heuristics implementation, training metamodels |
| **Intelligent technologies** | Rule-based inference, fuzzy logic | Classification and regression models | XAI/IML models, domain semantics |
| **Data processing** | Implicit, during knowledge acquisition | Data filtering and integration, model | Metamodel training |

|                  |                                          | training (identification and updating)                   |                                                          |
|------------------|------------------------------------------|----------------------------------------------------------|----------------------------------------------------------|
| **Data types**   | -                                        | Patients and clinical cases                              | Application of models, decision making                   |
| **Data integration** | -                                    | Yes                                                      | No (unified data)                                        |
| **Metamodeling** | Modification of knowledge base           | Tuning of model training, surrogate modeling             | Tuning of metamodel training, semantic knowledge update  |
| **Context**      | Clinical groups by official policies     | Subclinical (hierarchical) groups                        | Problem domain                                           |
| **Limitations**  | Diversity in clinical groups, uncertainty | Limited scalability, lack of systematic lineage to domain | Lack of systematic lineage to domain                    |
| **Validation**   | Evidence-based (during preparation), domain knowledge | Cross-validation, statistics                | Simulation-based, meta-model-based, experimental implementation |
| **Trust sources** | Official rules                          | Validation, explanation                                  | Interpretation, linking to semantics                     |

An important part of a holistic decision support system within the proposed framework is to narrow the context during the stages. Commonly, clinical recommendations, policies, and other sources of knowledge in Stage 1 are relatively broad, covering key healthcare regulations. Stage 2 is focused on working in subclinical groups by identification of classes, lowering diversity within a group, etc. Finally, Stage 3 deals with individual predictions, outliers, and explanations. Context control connects stages into a chain providing the right higher-level context for consecutive stages with the results of application of the previous stage(s). Building such a chain enhances possible automatic application, scaling, and extension of the solutions. Moreover, building a context-narrowing pipeline enables quantitative assessment and automation of value-based healthcare (VBHC) approach. VBHC considers diverse patient flow with requests for diverse care, even in a single clinical group. As a result, VBHC can work in all stages, while common evidence-based medicine is usually limited to Stage 1.

Additionally, the interconnection of stages enables mutual support to overcome limitations such as lack of validity and trust in particular stages since, in many cases, the available data and knowledge sources and validation procedures are located in different stages. So, the holistic framework interconnects and enhances each stage within a general pipeline for decision support.

### 3.4 Clinical background

Among many systematic reviews on the application of data-driven clinical decision support, only a few describe proven benefits for clinical outcomes [48]–[50]. Nevertheless, many projects have operated too small samples or lasted too little to reveal clinically significant results related to patient outcomes [51], [52]. However, we can see substantial evidence that CDSSs can certainly impact healthcare providers' efficiency, with preventive care reminder solutions and drug prescription services being the most explicit examples [53]. The experience related to diagnostic CDSS can be explained that doctors, based on their clinical experience, can better come up with alternative diagnoses and rule them out than a CDSS can [54]. In many cases, CDSSs are not able to detect and consider comorbid conditions, which is a big limitation to their use in real clinical settings. All these factors lower the impact of such CDSSs in clinical practice.

Furthermore, data-driven diagnostic models often require input or import of a large amount of patient data to provide the result of the diagnostics. As long as this information is not available in machine-readable form, e.g., in a semantically interoperable electronic medical record (EMR), clinicians have to enter missing data manually. The problem of manual data entry may make doctors give up or make mistakes, which can lead to the situation when CDSSs are not increasing the efficiency, but require a lot of time and effort for operation [55], [56].

Many modern solutions use automatic language processing and speech interfaces [57], [58] that facilitate data entry by clinicians. However, systematic reviews show equivocal evidence of improved processes and data quality [59]–[61].

To provide an efficient CDSS for real clinical settings, developers should study how to increase CDSS content to consider multiple comorbidities simultaneously, how to provide and to estimate the effect of a CDSS on clinical and organizational outcomes, and how CDSSs can be most effectively integrated into the workflow and deployed across diverse settings [54]. Clinical validation of CDSSs is a factor that can make their implementation successful. Providing the evidence of clinical efficiency is a resource-consuming task that requires support from a sustainable validation methodology [62][63].

## 4. Experimental implementation

To implement the proposed framework, a platform for model-based treatment support was developed. It includes both general instrumental solutions and basic methodological procedures for the application of the framework. Fig. 2 represents the general architecture of the platform, which includes three core blocks with implemented procedures.

1) The model training block includes a pipeline commonly implemented by data scientists. Still, the pipeline was extended in two aspects. Firstly, it explicitly follows the rule base constructed after the official guidelines, scales, recommendations, etc. to enable referring them within a context of data-driven model application. The rules are used in data selection, preprocessing, and filtering. Secondly, the automatic clinical pathway structuring and analysis [64] reveal actual clinical experience in a diverse patient flow even in a single disease. Such decomposition is further used to train a set of data models for subclinical groups and provide a trained model with meta-information for their application. This block was implemented as a series of Python scripts for training models in advance. Within the implementation, the trained models were stored in *.pickle files following the predefined interface and implementing methods for checking model applicability and applying it.

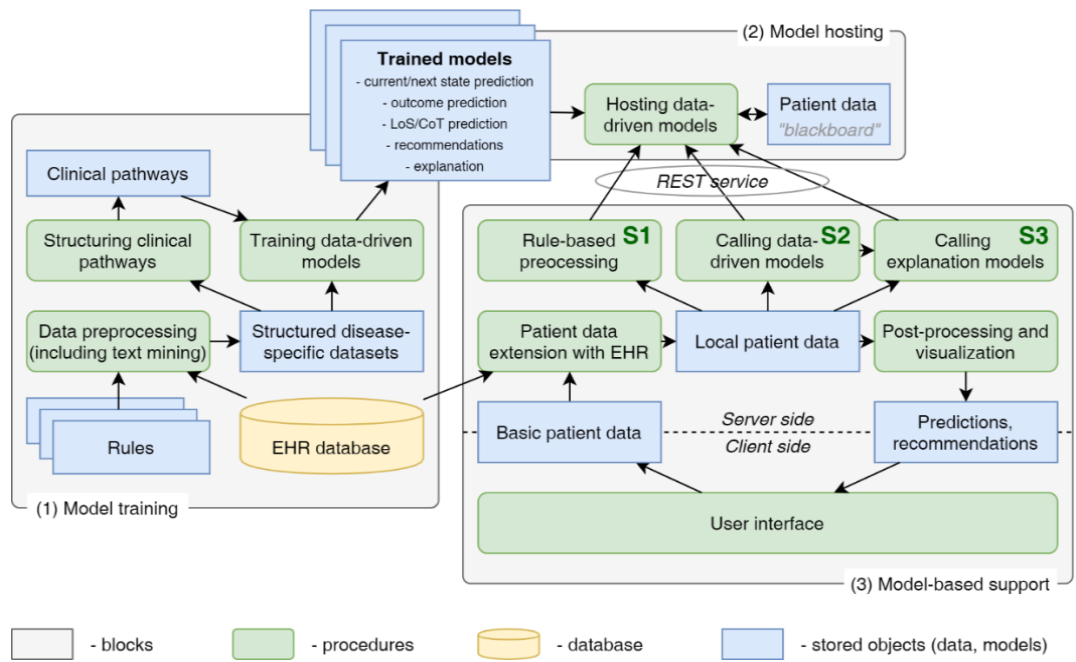

Figure 2 – General architecture

2) The model hosting block provides a service for accessing trained models. It was implemented in Python using Flask framework and a "blackboard" design pattern for applying available models to the dictionary of patient characteristics. The model-hosting procedures were implemented as a RESTful (Representational State Transfer) web-service accessible from various information systems (including mobile applications and medical information systems, MIS [65]). The blackboard pattern enables models from all three stages of the framework to be arranged using a unified interface. At the same time, implicit implementation of the framework in the hosting algorithm applies the models sequentially by stages.

3) Model-based support can be implemented in different MIS (e.g. in an extension module of such system) for physician support during treatment procedures. In addition, it could also be implemented in the form of personal assisting software for patients to assess their current health condition (including applications running on mobile phones, and many others).

The implemented blocks are loosely coupled. Block 1 generates models that are used in block 2 in serialized form. Blocks 2 and 3 interact via RESTful service. Finally, block 3 provides user access either directly through the dedicated UI or by extending an existing system (e.g., MIS). Within the framework, the user access is provided to a decision maker (e.g., to a physician deciding on the patient's treatment strategy).

During our study, we've been mainly focused on chronic diseases: arterial hypertension, diabetes mellitus, and chronic heart failure were selected as a running example. Each disease was considered within the scope of the proposed framework. At Stage 1, guidelines and scales applicable in the practice of physicians were considered and used for data preprocessing and baseline prediction. At Stage 2, predictive models were developed, trained, and applied based on prepared data and identified typical clinical pathways. Finally, at Stage 3, additional explanation and interpretation of models were applied for better clinical decision support. A summary with references to further reading is provided in Table 2. References for Stage 1 were provided to official clinical sources. References for Stages 2 and 3 were provided for a detailed description of our implementation of the models.

Table 2 – Implemented pipelines in chronic disease treatment support (overview, references)

| Disease | Stage 1 | Stage 2 | Stage 3 |
|---|---|---|---|
| **Diabetes mellitus** | FINDRISC, guidelines by ESC [67] | Complications, compensation status [68] | Personalized trajectories [68], [69], feature-based explanation |
| **Arterial hypertension** | SCORE [70], guidelines by ESC [71] | Therapy effectiveness prediction [72] | Controllability prediction [65] |
| **Chronic heart failure** | CHA2DS2-VASc [73] | Stage prediction [74] | Static and dynamic feature importance [74] |

## 5. Diabetes mellitus case study

This section provides a detailed explanation of a case study on the implementation of the proposed framework for a single disease, namely, type 2 diabetes mellitus (T2DM).

### 5.1 Case study methods

Each of the three stages of clinical decision support are shown using the example of solving a particular practical problem of creating a decision-making support system for the early T2DM diagnosis. In the first stage, we use the knowledge and experience of expert endocrinologists to select predictors: signals of insulin resistance. Also, together with experts, we analyze existing modeling methods to find the most suitable methods for the specifics of the task. In the second stage, we create predictive models estimating the future risk of developing T2DM. In the third stage, we use explainable artificial intelligence techniques to interpret modeling results. Further, we validate these results using common quality metrics, expert-based validations, and case-based surveys of physicians.

*Stage 1: Basic reasoning*

Diabetes mellitus (DM) is one of the most common chronic diseases in the world. Experts from the World Diabetes Federation predicted that the number of patients with diabetes will increase by 1.5 times by 2030 and will reach 552 million people. The increase happens mainly due to patients with T2DM [75]. For healthcare, this type of diabetes presents one of the highest priority problems, since this disease is associated with a large number of concomitant diseases, leading to early disability and increased cardiovascular risk. Studies have shown that early detection of patients prone to insulin resistance and timely preventive measures reduce the risk of developing the disease in the future. Therefore, special methods are needed to identify the patients at risk.

To date, several diabetic scoring algorithms are used in medical practice in various countries: AUSDRISK, DRS – Diabetes Risk Score, Omani Diabetes risk score, FINDRISK, Danish Diabetes Risk Score, and others [76]–[80]. Researchers describes several approaches to treating diabetic scoring scales. A distinctive feature of the first approach is the inclusion of ethnicity characteristics in the set of attributes for calculating T2DM risk. An example of this approach is the Australian type 2 diabetes risk assessment tool (AUSDRISK) scale. This scale includes smoking status, age, and gender. The second approach is to create a model for a single ethnic population. For example, the Omani Diabetes risk score was created to identify high risk among Omani Arabs. The scale showed good results for its population, but a decrease in the prognosis quality for other populations is probable. The third approach to building scales is based on including only those predictors that can be identified using the survey. This approach includes the Danish Diabetes Risk Score. The method includes age (30–60), gender, body mass index, the presence of arterial hypertension, physical activity, and genetic predisposition for diabetes. This is a high-quality method. However, it predicts the presence of diabetes at the time of measurement. It is impossible to assess the diabetes mellitus risk for a future long-time interval.

One of the most common in medical practice is the Finnish FINDRISK scale questionnaire. This questionnaire is used in medical practice in Russia. It was validated in the Russian population and showed good results [81]. This questionnaire was also tested on populations of other countries and showed a high quality of the prediction [82], [83]. However, this method assesses the risk over the next ten years. Risk assessment for a shorter time interval will allow better planning of preventive actions.

Therefore, it is necessary to create a new high-quality scale with a relatively short prediction interval. We selected features for calculating T2DM risk based on the scales mentioned above; studies of features that influence the risk of developing diabetes [79], [84]–[89]; current clinical guidelines [90], and expert opinion of endocrinologists from Almazov National Medical Research Center[1] (one of the leading cardiological centers in Russia). Below, we consider the advantages of the scales and try to deal with their shortcomings.

*Stage 2: Data-driven predictive modeling*

The study included 4,597 patients: 2,534 men and 2,063 women. Patients were divided into two subgroups: the first subgroup includes 90% of patients who did not show signs of insulin resistance for the following five years, the second group includes 10% of patients who revealed chronic type 2 diabetes mellitus five years after the observation (after measurements). All participants were treated at Almazov National Medical Research Center in 2000–2019.

---

[1] http://www.almazovcentre.ru/?lang=en

Criteria for patient entry into the study:
- Men and women without prediabetes and diabetes of any type and form (including gestational diabetes);
- Age from 18 to 90 years old;
- Observation period of at least five years;
- Lack of pre-diabetic and diabetic measurements of glucose and glycosylated hemoglobin at the beginning of the observation.

To develop a data-driven predictive model, a selection of machine learning models was used considering the following models: K-Neighbors Classifier, Random Forest Classifier, Logistic Regression, SVC (support vector classifier), XGB Classifier, Gaussian NB (Naive Bayes), Bernoulli NB, Multinomial NB. Parameters were selected using the cross-validation grid-search method. The final quality was checked on a test sample, i.e. data that wasn't used in model training. Feature engineering was sequentially reducing the initial set of features and checking for changes in the model's quality.

*Stage 3: Explaining and detailed fact reasoning*

Also, we compared the developed method with the one most widely applied in medical practice: the FINDRISK scale. Characteristics of the scale for comparison:
- Input: a vector of 8 medical indicators identified during the survey;
- Output: the probability of developing diabetes within ten years;
- Calculation Algorithm: logistic regression trained on precedents;
- Model accuracy: sensitivity of 76.0% and specificity of 60.2% for validation in the Siberian population (9360 people aged 45–69) [81].

**Physicians perceiving**

To assess the practical effect that CDSS applied in the real pipelines we have conducted a survey with physicians working with corresponding nosology (T2DM). The survey was structured into two phases. The first phase is based on the case-based assessment of model prediction within three sets, namely, a) with model only, b) with FINDRISK scale presence, c) with FINDRISK scale and explanation presence. The second phase is based on a general system assessment after the first phase. The study was aimed at investigating physicians' acceptance level as well as the influence of different factors on this level. Mainly the structure of the study is based on the theory of planned behavior (TPB) [91], [92] considering attitude, subjective norms, perceived behavioral control influencing behavioral intentions (and actual behavior). Within the current study, we consider confidence and understanding of the model as core characteristics of the CDSS perception (similarly to [93]). As such, we introduce agreement and understanding of the information as additional performance characteristics in addition to intention to use (furtherly denoted by "Agree", "Understand", "Use" characteristics). Also, we consider a relationship a) between subjective norms and integration with accepted practice (at Stage 1), b) between attitude (and future behavior control) and explainability, validity, integrity of the model (at Stage 3). We consider "Agree" as a first criterion reflecting general user's attitude, "Understand" as a second one reflecting correspondence to subjective norms and perceived control, "Use" as a final target criterion reflecting obtained intention to use the solution.

**Physicians recruitment**

Physicians were recruited from the inpatient and outpatient clinics in Saint-Petersburg, Russia. All of the physicians did not have experience using the decision support system by the moment the study began. All the study participants signed an informed consent that clarified the goals of the study and its procedures including data privacy and security measures. The study included 19 physicians with the following details: 14 physicians were from outpatient practice (the other 5 works with inpatient), average work experience is 8 years. The physicians had the following specializations: therapist (10 doctors), endocrinologist (3 doctors), cardiologist (2 doctors), gynecologist (2 doctors), ophthalmologist (2 doctors).

**Case assessing**

This phase includes a questionnaire based on random anonymized retrospective T2DM cases from Almazov national medical research center database. Each case was presented to the physician (see Fig. 3) with the patient's basic information (anamnesis, sex, age, height, weight, BMI (body mass index), antihypertensive therapy, physical activity level, blood sugar, heredity, arterial pressure) and one of three prediction settings (randomly selected). The prediction model settings were as follows:

− Setting A: estimation of T2DM risk according to the model (Stage 2 only);
− Setting B: same as above and estimation of T2DM risk according to the FINDRISK scale (Stage 1 and 2);
− Setting C: same as above and estimation SHAP-based explanation of the model prediction (all stages).

The physicians were asked to assess each case with three scoring variables ("Agree", "Understand", "Use"). To rate each item, a Likert scale from 1 (strongly disagree) to 5 (strongly agree) was applied.

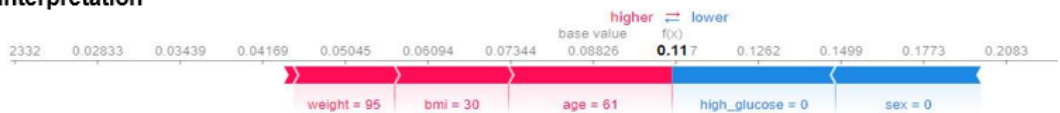

Figure 3 – Testing interface (Setting C)

To investigate the obtained phenomenon furtherly, we performed an analysis of interconnections between scoring and selected parameters. In addition to the presented scoring, we introduced variables showing risk assessment by the FINDRISK scale and by the model treated as categorical variables with three values "low", "mid", "hi". Also, the presence of the FINDRISK scale and explanation in the setting were considered as binary variables. The variables were normalized with corresponding scales to interval [0,1]. Furtherly, mutual information ($MI$) was estimated for available variables. The interconnections between variables were selected with the threshold $MI > 0.03$ (selected to keep the variable graph connected). The direction for causal assessment was selected according to the following assumptions: a) the risk and presence variables can influence scoring variables, b) the scoring variables were ordered ("Agree" → "Understand" → "Use") to reflect TPB approach treating "Use" as behavioral intension. Next, linear regression models were obtained for each node having inbound connections.

**System assessing**

After the first phase, we invited the physicians to participate in the study of the user acceptance of the system. The study participants were interviewed in person by one of the research team members. The invitation was accepted by 16 physicians (the remaining 3 physicians were not able to participate the second phase of the survey due to the various personal reasons).

To assess the user acceptance of the system, a Wilson and Lankton's model of patients' acceptance of electronic health solutions was applied [94]. The model allowed measuring the following criteria: behavioral intention (BI) to use, intrinsic motivation (IM), perceived ease-of-use (PEOU), and perceived usefulness (PU) of the decision support system. BI represents the intention to utilize the system and to rely on the decision support that it provides; IM represents the willingness to use the system provided that no direct compensation is available; PEOU represents the extent to which the provided reports are clearly presented and comprehended by users; and PU denotes the degree to which the doctors believe that the utilization of the decision support system will improve their experience with patient risk assessment. BI measure consisted of 2 objects, whereas IM, PEOU, and PU consisted of 3 objects each. Russian translation of the questionnaires made by the research team was used during the study. To rate each item, a Likert scale was applied.

Table 3 – User questionnaire: user acceptance evaluation

| |
|---|
| **1. Behavioral intention to use**<br>　　a. I intend to use the tool to have a second opinion on the patient risks<br>　　b. I feel like I will use it in the future<br>**2. Intrinsic motivation**<br>　　a. The system helps me to make more informed decisions<br>　　b. I trust the system as it provides interpretations of the results<br>　　c. I trust the system because it provides references to the standard scales<br>**3. Perceived ease of use**<br>　　a. The model outcomes are clear and understandable<br>　　b. The interpretations are clear, and I understand the reasoning<br>　　c. The visualizations are clear and I don't spend much time on their interpretation<br>**4. Perceived usefulness**<br>　　a. Using the system enhances the effectiveness of managing risks of my patients<br>　　b. It explains me why a certain risk assessment is done<br>　　c. I can provide all the information about my test results to any doctor I visit |

All the study participants were given individual access to the online questionnaire (Table 3), which they were asked to fill in. All the participants received a written detailed instruction on how to operate with a questionnaire and the sense of the rating scale.

After we collected and analyzed the results of the user acceptance evaluation, we have organized a study to deeper understand the reasoning of the doctors when working with the CDSS. The study was made with semi-structured one-to-one expert interviews based on an interview topic guide [95]. A semi-structured topic guide was developed by the study team (Table 4). The interview scripts were recorded and analyzed later by every member of the research team.

Table 4 – User questionnaire: interview

| |
|---|
| 5. Can you understand a model output without interpretations?<br>6. Are you convinced with the interpretations that the system provides?<br>7. Do you require an interpretation to be able to critically assess a model output?<br>8. Does a reference to a standard scale facilitates critical assessment of a recommendation?<br>9. Could you provide any comments/suggestions that, in your opinion could improve the CDSS, if any? |

**Human subjects and privacy protection**

All the experiments involved only anonymized retrospective cases. No patients were physically involved during the experimental study. Additional manual check of the cases presented to the physicians during the study was performed to ensure there was no private patient data exposed. No private data or data enabling patient identification were found during the check. With the mentioned conditions, the study did not require approval by an ethics committee.

**5.2 Case study results and discussion**

**Predictive modeling**

As a result, we obtained a model for assessing the risk of type 2 diabetes over the next five years.

Characteristics of the final model:

− Required software: Python 3.6;
− Input: a vector of 6 medical indicators influencing the risk of T2DM;
− Output: the probability of developing T2DM within the next five years from the moment of the last measurement;
− Calculation algorithm: gradient boosting over decision trees trained on precedents;
− Model accuracy: accuracy of 70% on the test samples (88% on the training samples).

Compared with the FINDRISK scale, the developed model works on the smaller number of features (6 and 8). The model does not require sophisticated medical tests. With similar quality of prediction (the sensitivity of 76.0% and specificity of 60.2% versus the sensitivity of 82.0% and specificity of 62%), the lead time of the model is better (5 years against 10). Therefore, our method is interpretable and scalable, and the model can be used to assess the risk of T2DM.

**Explaining and reasoning**

As an example of the model's interpretation, we look at SHAP values (Fig. 4) for the top three most important features. The features of body mass index at medium and high values increase the probability of developing T2DM by an average of 0.4. The mean diastolic blood pressure (DBP) increases the probability of T2DM development by an average of 0.2 at high and medium pressure values and decreases it by an average of 1.2 at low values. The body surface area contributes to prediction of increased risk only at high values, with average values the average contribution is close to 0, while low values indicate reduced risk.

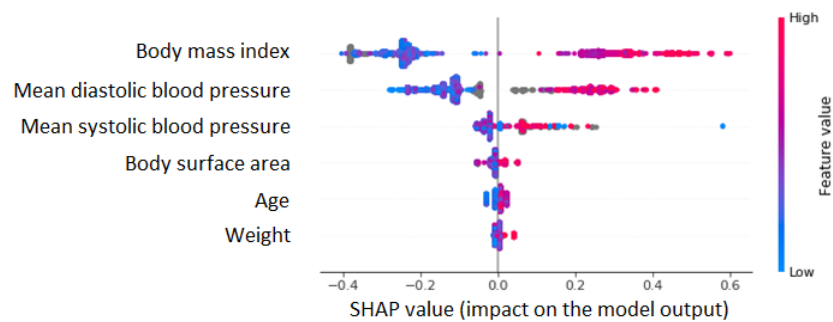

Figure 4 – SHAP chart for the model

We demonstrate an example of the proposed model's application using treatment cases of two patients. The first patient is a man, born in 1957, without signals of insulin resistance at the beginning of observation. The second patient is a man, born in 1951, with no signs of insulin resistance in medical history too. The vectors describing conditions of these patients at the beginning of observation are 31/68/55/116/87 / 2.007 (patient #1) and 33/82/56/128/109 / 2.38 (patient #2) for signs BMI / mean_dbp (mean diastolic blood pressure) / age / mean_sbp (mean systolic blood pressure) / weight / BSA (body surface area) respectively.

Using these vectors, the proposed model estimates the risk of developing diabetes over the next five years as 39% for the first patient and 60% for the second one.

Fig. 5 show the contributions of each characteristic to the prediction, contributions are calculated using the SHAP method.

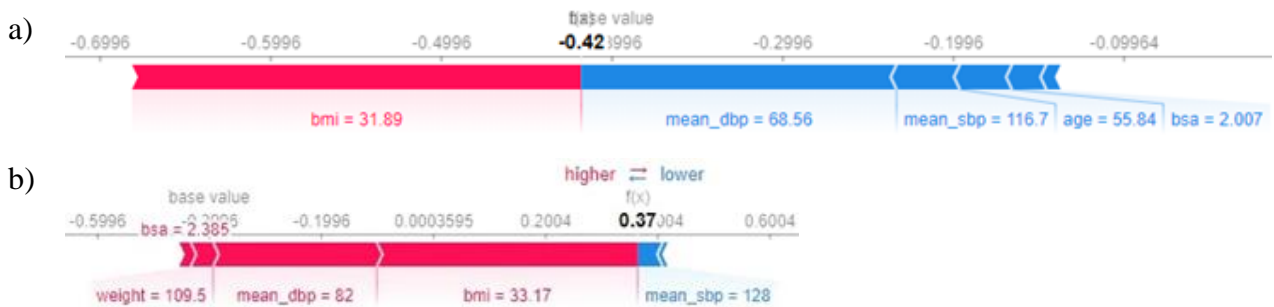

Figure 5 – Contributions of each indicator to prediction examples a) patient #1, b) patient #2

For the first patient, a low systolic blood pressure combined with a low mean diastolic blood pressure greatly reduced the estimate of developing the risk of diabetes, overcoming the contribution of BMI 31.89 (grade 1 obesity) to increasing the estimate of diabetes risk (Fig. 5a). For the second patient, the systolic blood pressure, BMI, weight, and BSA are above normal, increasing the estimate of diabetes risk. The mean systolic pressure is within the normal range and thus reduces the risk. However, its contribution is less than the total contribution of the other indicators towards increased risk (Fig. 5b). Thus, the prediction is the absence of developing diabetes for the first patient and the presence of developing disease in the observation period for the second patient. Predicted values have coincided with the real observed values for both patients. Diabetes did not develop during the observation period for the first patient and was diagnosed 4 years and 11 months after the start of the observation for second patient. In this example, we have shown that use of the proposed framework can result in creation of interpretable, usable, and high-quality models.

People with T2DM are often asymptomatic until complications develop. For this reason, T2DM may remain undetected for many years. Moreover, during the so-called pre-diabetic stage, when people have either impaired glucose tolerance or impaired fasting glucose, damage to blood vessels and nerves may already be underway [96]. Thus, the detection of individuals with prediabetes and undiagnosed T2DM is an important approach to prevent or delay the development of T2DM or its complications. To date, a lot of prediction methods have been developed to identify individuals with unknown T2DM or those with a high risk of T2DM development. Among all predictive modeling approaches, machine learning methods often achieve the highest prediction accuracy. Still, most machine learning models do not explain their prediction result, which precludes their widespread use in healthcare [97]. As was reported in a systematic review in 2013, of the 65 non-invasive diabetes risk assessment tools available worldwide, only ten have reported on their use as a screening tool [98]. Findings from this review suggested that diabetes risk assessment tools were not widely used in practice. The barriers preventing healthcare practitioners from implementing diabetes risk assessment tools were their attitudes toward the tools, impracticality of using the tools, and lack of reimbursement and regulatory support. [98] Also, considering the differences in ethnic specificity and influencing factors in different regions, the majority of prediction models may not be readily applicable to all people worldwide. Prediction tools always need to be validated for each population before implementation [99]. The model developed in this study predicts the probability of T2DM development in the next five years, has reasonable accuracy, sensitivity, and specificity (70.0%, 76.0%, and 60.2%, respectively), is non-invasive and simple to use, includes only six variables (age, weight, body surface area, body mass index, mean systolic and diastolic blood pressure). These six variables are easy to obtain in clinical practice. Commonly, electronic patient records contain complete information required for this prediction tool, so it can be used to assess an individual's risk without direct face-to-face interaction. Thus, this model can be implemented for detecting individuals with a high risk of T2DM at the population level. However, such variables as blood pressure and body mass index may require an assessment by healthcare providers, so this model may be hard to use as a self-administered tool (by patients). Another limitation of the model is the fact that all patients included in the study were from one medical center. The model's accuracy may change when tested in different cohorts.

For this reason, the tool needs to be tested and validated in other studies. Concerning the interpretability of the model, it is provided with clear, easy-to-understand plots providing basic explanations of how the model works. From these plots, it becomes clear how much a particular factor contributed to the prediction at a particular value (SHAP plot) and at what value of the factor the highest prognostic capability is achieved (Partial Dependence Plots).

In summary, the T2DM prediction model developed in this study can be proposed to estimate the risk of T2DM in routine clinical practice after testing and validation on other samples. Based on the results of T2DM risk assessment by this tool, the following recommendations can be made: further screening (invasive blood glucose tests such as fasting plasma glucose levels or oral glucose tolerance tests) for patients with high risk and lifestyle modification to prevent diabetes in the long term for patients with moderate risk.

**Physicians perception**

During the study, we have collected 570 answers with approximately equal numbers of responses for different settings (169, 200, 173 for A, B, C, respectively). On average each physician reviewed 30 cases

The physicians disagree (score 1 or 2) with the model in 13% of cases. In those cases, the physician claims that the model shows lower risk in 42% of cases and shows higher risk in 58% of cases. Also, in the disagreed cases, the validation (with the existing case behind the anonymized EHR) shows that model is correct in 66% of cases, while the physician is correct in 34% of cases.

Table 5 – Case scoring mean (95% confidence interval)

|  | **Agree** | **Understand** | **Use** |
|---|---|---|---|
| **All settings** | 3.93 (3.84, 4.02) | 4.21 (4.12, 4.31) | 3.93 (3.84, 4.02) |
| **Setting A** | 4.11 (3.94, 4.28) | 4.40 (4.23, 4.56) | 4.08 (3.91, 4.25) |
| **Setting B** | 3.75 (3.59, 3.91) | 4.16 (3.99, 4.32) | 3.87 (3.72, 4.02) |
| **Setting C** | 3.90 (3.73, 4.07) | 4.20 (4.03, 4.36) | 3.87 (3.70, 4.04) |

The average scoring is shown in Table 5 with a 95% confidence interval. During the analysis, we discovered the highest scores were obtained in Setting A, while the lowest is obtained in Setting B. Although it was an unexpected result (the supposed order initially was C-B-A), after the discussion with the physicians and further investigation, we consider that the obtained order is the result of the "criticality" level available for a physician faced with the prediction results. Without the presence of the standard scale, a physician trust blindly the model results that increasing number of type I errors, which can be lowered in comparison to the basic scales. At the same time, the presence of an explanation increases all scores showing the importance of Stage 3 within our framework.

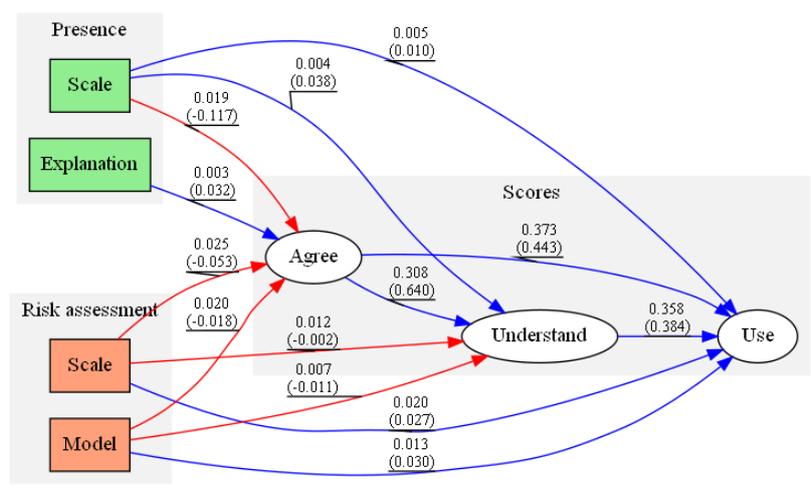

Figure 6 – Interconnection between case scoring and variables

Fig. 6 shows the resulting graph with *MI* and the linear coefficient shown for each edge (shown in brackets). We see that the high risk according to both the model and the FINDRISK scale have a negative influence on "Agree" (more) and "Understand" (less) scores which to our believes reflect the fact that the more complicated the case is the more questionable the predictions are. Also, we see that the presence of the FINDRISK scale has negative influence on "Agree" score. That matches the idea of physician's "criticality" increasing with the presence of a standard scale and high patient risk according to the models. At the same time, high risk according to both scales positively influence "Use" scorers that can be considered as evidence of model importance for complex cases. In addition, the presence of explanation and standard scale positively influences the "Agree" and "Understand" scores respectively. Different directions of positive influence of these two elements may be considered as evidence of importance of both Stage 1 and Stage 3 within the proposed approach.

**User acceptance**

The median values for BI, IM, PEOU, and PU (3, 3, 4 and 4 respectively) showed that the users generally accepted the decision support system and the reports that it generates (Table 6).

Table 6 – Acceptance criteria

| **Criterion, Item** | **Median** | **Max** | **Min** |
|---|---|---|---|
| **1. Behavioral intention to use** | 3 | 5 | 2 |
| 1a. I intend to use the tool to have a second opinion on the patient risks | 3 | 5 | 3 |
| 1b. I feel like I will use it in the future | 3 | 4 | 2 |
| **2. Intrinsic motivation** | 3 | 4 | 2 |
| 2a. The system helps me to make more informed decisions | 3 | 4 | 2 |
| 2b. I trust the system as it provides interpretations of the results | 3 | 4 | 3 |
| 2c. I trust the system because it provides references to the standard scales | 3 | 4 | 2 |
| **3. Perceived ease of use** | 4 | 5 | 2 |
| 3a. The model outcomes are clear and understandable | 4 | 5 | 3 |
| 3b. The interpretations are clear, and I understand the reasoning | 4 | 4 | 2 |
| 3c. The visualizations are clear and I don't spend much time on their interpretation | 4 | 5 | 3 |
| **4. Perceived usefulness** | 4 | 5 | 2 |
| 4a. Using the system enhances the effectiveness of managing risks of my patients | 4 | 4 | 2 |
| 4b. It explains me why the a certain risk assessment is done | 4 | 5 | 3 |
| 4c. It explains me why the a certain risk assessment is done | 4 | 5 | 3 |

Minimum rates of 2 show that some users are still cautious towards the system. As all Median rates were in the positive part of the scale, we can see a general acceptance and encouragement to use the CDSS. This also indicates general acceptance of the solution by the doctors. They intend to use the system especially as a second opinion (median 3 out of 5). Doctors say that their motivation is supported by the ability of the system to provide references to the standard scales and interpretations (median 3 out of 5). The system is easy to use (median 4 out of 5). It provides clear and understandable outcomes and reasonings that are well visualized (median 4 out of 5). The system is useful for the doctors, they say that it helps to manage risks (median 4 out of 5). All the criteria showed a median of 3 to 4, which indicates a good level of user acceptance of the system.

We have analyzed the answers of the participating doctors to understand the reasoning behind the acceptance evaluation. Selected answers are presented in the Table 7.

Table 7 – Selected interview answers

| | |
|---|---|
| 5. Can you understand a model output without interpretations? | Without interpretation, it is not possible to understand why the model determined this risk for this patient [A1] |
| | Interpretation algorithms allow us to see which predictors in a particular patient increase the risk of pathology to the greatest extent [A2] |
| 6. Are you convinced with the interpretations that the system provides? | They are quite convincing in most cases. But there are rare cases when the graph built to interpret the model shows dependencies that are not logical and clinically explainable [A3] |
| | Shows well what factors matter most, and everything makes sense [A4] |
| | Pretty convincing, but for patients the risk as a percentage means roughly nothing, it makes no difference to them whether the risk is 20 or 30%, what matters is what exactly needs to be done and why [A5] |
| 7. Does a reference to a standard scale facilitates critical assessment of a recommendation? | This is necessary, for example, to see that the model has calculated too high a risk because of a high value of some factor which, in fact, unlike other patients, is not so important in this patient for some reason [A6] |
| | This is to understand what risk factors to look for first in this patient [A7] |
| 8. Does a reference to a standard scale facilitates critical assessment of a recommendation? | While we understand that this is a well-tested tool in many studies, I, for example, think that it may lack the data to properly assess risk [A8] |
| | In practice, I think yes, it's good that there's a reference to a standard scale for calculating risk [A9] |
| | Yes, because the model does not take into account research data and clinical guidelines [A10] |
| 9. Could you provide any comments/suggestions that, in your opinion could improve the CDSS, if any? | In my opinion, the model should still be defined by or with a doctor [A11] |
| | Since the tool can be used by patients, for example, as part of health screenings, it would be useful to translate the results from medical to "common" Russian [A12] |

From the answers that the doctors provided, we can conclude that interpretability plays an important role in understanding and accepting a CDSS output ([A1]), especially when interpretation is done on the feature basis [A2]. The system's output is convincing, and the doctors can act based on it. Interpretations also help doctors to identify incorrect conclusions when a system produces them [A3-A4]. They still see room for improvements, as not everything should be measured in numbers [A5]. The doctors see the importance of combining data-driven output with rule-based scales [A6-A7]. Despite the understanding that data-driven models are based on high-quality, real-world data [A8], doctors still ask for standard and known tools [A9] as they are based on the research and clinical guidelines [A10]. Doctors still believe that experts should be involved in the model development [A11]. This can potentially help to expose the results of the CDSS even to patients [A12].

An important part of the survey was feedback from the physicians within the answers on questions 5-9. Most of the physicians outlined the importance of feature-based explanation for model result understanding. Also, mentioning the importance of introduced steps (normative references at Stage 1, explanation at Stage 3), the physicians suggested extending the supporting elements requesting more features, more normative references like official clinical recommendations. An important aspect mentioned within feedback was the high speed of study without explicit expert involvement during the CDSS development. One of the possible directions was a patient-aimed recommendations variant with fewer details but with suggested steps to be taken (e.g., lifestyle recommendations, diet, etc.).

## 6. Discussion and future work

We have applied the proposed framework to the analysis and prediction of T2DM, where multiple scales are used in clinical practice [76]–[80]. The framework was elaborated through all three stages, and the results of data-driven predictive modeling were interpreted with both quantitative and structural comparison to the FINDRISK scale. This enabled a conceptual loop over all three stages and provided the following functionality:

1) The three-stage framework enables validation of data-driven models in comparison with the existing (trustable, manually checked, and widely applied in practice) scoring algorithms. This validation extends a common data-driven modeling pipeline with additional capabilities to assess the quantitative performance of the models (compared to the scoring algorithm) and their structure (feature engineering compared to the feature selection in the scoring algorithm).

2) The data-driven procedures mapped onto an interface similar to that of the existing scoring algorithm enables natural extension and modification of the algorithm based on its application results during the improvement cycle. This methodology provides a hybrid manual scoring vs. data-driven concurrency environment where continuous improvement could be available through modification of knowledge in Stage 1.

3) Further, the implemented loop can be considered as a process of coevolution of the whole three-stage pipeline. It can improve the overall performance and functionality, together with integrability and interpretability of the existing clinical BPs. Conceptual integration of all stages provides a human-understandable interface for DSS that can naturally work in practice.

4) The introduced additional stages increase the "criticality" of physician's view of the model results. As a result, whether a model-based prediction was accepted, a physician is more confident with the decision and trusts the model with the presence of references to standards (scales, clinical recommendations) and explanation. Also, we believe that the phenomenon of "criticality" is a worth much deeper investigation to increase trust and validity of CDSS in real systems, which include also social and informational aspects. Thus, we consider this investigation as one of the future directions in the development of our framework.

The proposed framework can be applied in a wide variety of contexts. Moreover, the framework can further be translated into multiple areas where legal regulation is combined with a large amount of data available for data-driven predictive modeling. Several examples of such areas may include law, education, human resource management, and many others. At the same time, the framework may support revealing and elaboration of contradictions between formal regulations (considered as an ideal situation) and practice (exposed within the available real-world data). Here, a contradiction may be discovered at the level of diversity description (while comparing and integrating Stages 1 and 2) and in real-word cases deviating from regulatory rules (while comparing and integrating Stages 1 and 3). Both may lead to the update of Stage 1 rules, e.g., reworking official recommendations. Furthermore, we believe that with a certain level of trust achieved by data-driven models, such models could be explicitly included in official rules, thus providing an enhanced level of automation and intelligent support of operational activity in the corresponding areas. Within the proposed approach, we introduced the developed predictive model as functionally similar to known and widely accepted scale (FINDRISK) to let the physicians compare and see the differences in performance and explanation (within second and third stages, respectively). The results are supported not only by the prognostic power of the developed models but also by the user acceptance analysis study made with practicing clinicians.

Our approach has a conceptual novelty, which makes it hard to find another approach to compare with. Still, it helps to overcome several of the main known barriers of CDSS adoption [100], [101], which are valid both for knowledge-based and data-driven CDSS. Among these barriers are the followings: a lack of knowledge provided about the decision and its context, lack of trust of the physician towards the system, lack of interpretation and decision back-tracing. Overcoming these barriers can facilitate the implementation of CDSS and raise an adoption rate among physicians, which was shown by the case study.

One of the possible directions of development is focusing on the real-world behavior of agents, including decision-makers (physicians), clients (patients), and others. For instance, the extension of the decision beyond simple utility expectations with behavioral or information-processing approaches became one of the most recent directions of the research [102]. One of the important aspects of the proposed framework is increasing awareness and understanding of the modelling results by a physician. We consider further development of behavior aspects (including criticality, acceptability, intentions and other cognitive characteristics) interconnected with the framework as important direction in our future work.

The enhanced scalability of the system can be achieved by the integration of the decision support system with MIS. As one of the further developments, we are working on implementing an integration module that supports CDS hooks [103]. So, the system will be able to work with FHIR-enabled hospital information systems [104]. Such integration will enable natural interconnection with the semantics of various decision contexts, diseases, and roles of decision-makers to support the system's scalability automation.

The issue of integration of the decision support functionality into BPs is quite crucial due to several circumstances. First, the existing BPs usually follow Stage 1 rules as it is prescribed to practitioners to work in a predefined way. A modification of such processes should consider limitations to provide proper and useful support. Second, any modifications should be applied to the existing BPs "as is" considering the human-centered activity of practitioners (not too ideal BPs). Third, a modification of a BP should consider all types of the system's elements, namely agents, IT solutions, resources, facilities, etc. Moreover, a DSS (and CDSS in particular) is an element of the IT sub-system. Therefore, good decision support should consider the peculiarities of human-computer interaction [3] and adaptation of support during the practice where both the DSS and BP are co-evolving to some stable state. We consider a further technological development and implementation of the framework in various scenarios and systems as another important direction for future work. This includes integration with MIS, development of patient-aimed solutions (mentioned in physicians' feedback after the survey).

## 7. Conclusion

In this paper, we have proposed a framework, which is based on a three-stage approach aimed to enrich the applicability of CDSSs and acceptance of them by physicians in daily practice. The framework is based on accomplishing DD-models implemented within a CDSS with explicit references to the existing clinical norms (scales, clinical recommendations) and domain-specific explanation of the results. The framework introduces best practices for building a CDSS that can be adopted within clinical pipelines, enhance the decision-making processes, and retain their supportive efficiency for a long period of time. The performed experimental study and following survey show the framework implementation within the T2DM treatment process and give promising insights on the framework applicability and future development directions. We believe that such extension is essential for delivering applicable and user-acceptable CDSS in clinical practice.

**Acknowledgments.** This work was financially supported by the Government of the Russian Federation through the ITMO fellowship and professorship program. This research is financially supported by The Russian Science Foundation, Agreement #19-11-00326.